\documentclass[copyright,creativecommons]{eptcs}
\providecommand{\event}{ACL2-2023} 
\usepackage{breakurl}             
\usepackage{underscore}           
\usepackage{float}

\usepackage{amsmath,amssymb,amsfonts,amsthm}
\usepackage{fancyvrb}
\usepackage{alltt}
\usepackage{color}
\usepackage{xspace}
\usepackage{listings}
\floatstyle{boxed}
\restylefloat{figure}

\ifx\BRRTEXT t
\newcommand{\hrefdocx}[2]{{}{#2}}
\newcommand{\hrefdoc}[2]{\hrefdocx{}{{\underline{#2}}}}
\newcommand{\hrefdoctt}[2]{\hrefdoc{}{\texttt{#2}}}
\newcommand{\hrefdoccl}[2]{\hrefdocx{}{{\underline{#2}}}}
\newcommand{\hrefdoccltt}[2]{\hrefdoccl{}{\texttt{#2}}}
\else
\newcommand{\hrefdocx}[2]{\href{http://www.cs.utexas.edu/users/moore/acl2/manuals/current/manual/index.html?topic=#1}{#2}}
\newcommand{\hrefdoc}[2]{\hrefdocx{ACL2\_\_\_\_#1}{{\underline{#2}}}}
\newcommand{\hrefdoctt}[2]{\hrefdoc{#1}{\texttt{#2}}}
\newcommand{\hrefdoccl}[2]{\hrefdocx{COMMON-LISP\_\_\_\_#1}{{\underline{#2}}}}
\newcommand{\hrefdoccltt}[2]{\hrefdoccl{#1}{\texttt{#2}}}
\fi

\ifx\BRRCOMMENTS n
\newcommand{\mycomment}[1]{}
\else
\newcommand{\mycomment}[1]{{\vspace{.1in}\noindent\color{red}{\bf{*Comment*: }\em{#1}\vspace{.1in}}}}
\fi

\newcommand{\nil}{\texttt{nil}\xspace}

\title{Advances in ACL2 Proof Debugging Tools\footnote{Released under Distribution Statement ``A'' (Approved for Public
Release, Distribution Unlimited).}
}
\author{Matt Kaufmann and J Strother Moore
\institute{Department of Computer Science,
The University of Texas at Austin, Austin, TX, USA
(retired)}
\email{\{kaufmann,moore\}@cs.utexas.edu}
}
\def\titlerunning{ACL2 Proof Debugging Tools}
\def\authorrunning{M.~Kaufmann \& J S.~Moore}

\begin{document}
\maketitle

\begin{abstract}

The experience of an ACL2 user generally includes many failed proof
attempts.  A key to successful use of the ACL2 prover is the effective
use of tools to debug those failures.  We focus on changes made after
ACL2 Version 8.5: the improved break-rewrite utility and the new
utility, \texttt{with-brr-data}.


\end{abstract}

\section{Introduction}

The key technique for debugging ACL2 proofs is known as
\hrefdoc{THE-METHOD}{{\em the method}}.~\footnote{The online version
  of this paper provides underlined links like this to
  \hrefdoccl{DOCUMENTATION}{documentation}~\cite{acl2-plus-books-manual}
  topics.}  Briefly put, it is to look at {\em checkpoints} that the
prover cannot further simplify, usually to get ideas for controlling
rewriting by introducing new rewrite rules or by
\hrefdoc{ENABLE}{enabling} or \hrefdoc{DISABLE}{disabling} existing
ones.  However, there are many proof debugging tools that can be
helpful when using the method; brief summaries may be found in the
list of \hrefdoc{DEBUGGING}{debugging} topics (for proofs and
otherwise) in the ACL2+books manual~\cite{acl2-plus-books-manual}.
Among these are a number of popular tools, including
\hrefdoctt{ACCUMULATED-PERSISTENCE}{accumulated-persistence} and the
\hrefdoc{PROOF-BUILDER}{proof-builder}.

However, this paper focuses on the following two tools:
\hrefdoctt{WITH-BRR-DATA}{With-brr-data} and the
\hrefdoc{BREAK-REWRITE}{break-rewrite} utility.  These are tools that directly
track the ACL2 rewriter.  The former was introduced after the release
of ACL2 Version 8.5 in July, 2022.  The latter was introduced in the early 1990s (Version 1.3)
but was significantly improved after the Version 8.5 release.  The
sections below deal with each of these, first at the user level and
then with implementation-level discussions.  The user-level
discussions start with the new tool, \texttt{with-brr-data}, followed
by a section on break-rewrite (including its new ``near misses''
capability) and then a section showing their use together.  The
implementation-level sections start with background material on a key
enabling device, \hrefdoc{WORMHOLE}{wormhole}s, followed by a
discussion of break-rewrite implementation, after which we discuss the
implementation of \texttt{with-brr-data} and how it takes advantage of
the break-rewrite implementation.  Next we discuss how to change the
functionality of \texttt{with-brr-data} with
\hrefdoc{DEFATTACH}{attachments}.  We wrap up with a conclusion.  We
assume some familiarity with
\hrefdoc{INTRODUCTION-TO-REWRITE-RULES-PART-1}{rewriting in ACL2}.

The examples in this paper are available in supporting materials for
this paper; see \hrefdoc{COMMUNITY-BOOKS}{community books} file
\texttt{workshops/\-2023/\-kaufmann\--moore/README}.  Additional
examples may be found in \texttt{demos/\-brr\--test\--input.lsp},
which generates output found in \texttt{demos/\-brr\--test\--log.txt},
and in \texttt{system/\-tests/\-brr\--data\--input.lsp}, which
generates output found in
\texttt{system/\-tests/\-brr\--data\--log.txt}.

\textbf{Terminology}.  An {\em application} $A$ of a rewrite rule $R$
is the process of replacing a term, the {\em target}, by suitably
rewriting the right-hand side of $R$ to obtain the {\em result} of
$A$.  This process includes not only rewriting the right-hand side of
$R$ but also relieving the hypotheses of $R$.
For this purpose, the equality $(f x_1 \ldots x_k)=b$ representing
a definition is viewed as a rewrite rule whose left- and right-hand
sides are those of the equality.  A term $t_1$ is said to {\em
  contain} $t_2$ if $t_2$ occurs as a subterm of $t_1$.  A rewrite
rule application {\em introduces $tm$ as a subterm} if the result, but
not the target, contains $tm$.  For a given application $A$ of a
rewrite rule $R$, a {\em subsidiary
  application} is a rewrite rule application that takes place after
$A$ begins but before $A$ completes; that is, it takes place during
the process of rewriting the right-hand side of $R$.  Thus when $A$ is
represented by a frame displayed by the utility
\hrefdoc{CW-GSTACK}{cw-gstack} or (similarly) the break-rewrite
\hrefdoc{BRR-COMMANDS}{\texttt{path} command}, all rewrite rule applications
represented below that frame are subsidiary to $A$.  Finally, a {\em
  top-level} operation of the simplifier is any operation performed by
the simplifier that is not made while attempting to relieve a
hypothesis.

\section{User-level introduction to
  \hrefdoctt{WITH-BRR-DATA}{with-brr-data}}

ACL2 users sometimes find a surprising term in a checkpoint.  This
section summarizes how the \texttt{with\--brr\--data} utility may be
used to see how rewrite rules were applied to produce a given term
during a proof attempt.  These rules might well include at least one
rule that hasn't occurred to the user, perhaps because it was
introduced by including someone else's book.  See also the
\hrefdoctt{WITH-BRR-DATA}{with-brr-data} documentation for more
details and examples.

The behavior of associated queries is perhaps best explained with the
examples below, but we start here with a summary.  If $F$ is a form
that invokes the prover, then \texttt{(with-brr-data $F$)} saves
prover data that can be queried later.  A key query is of the form
\texttt{(cw-gstack-for-subterm $tm$)} where $tm$ is a term; variants
are described near the end of this section.  That query searches for
the first top-level rewrite rule application, $A$ --- the {\em
  product} of the search --- that introduced $tm$ as a subterm.  If
$A$ is found, then a stack $S$ is displayed as with
\hrefdoctt{CW-GSTACK}{cw-gstack} or (similarly) the break-rewrite
\hrefdoc{BRR-COMMANDS}{\texttt{path} command} (hence using terms in
\hrefdoc{TERM}{translated} form).  $S$ represents top-level operations
down to a frame $F$ that represents $A$, and $S$ maximally extends past
$F$ such that every rule application represented below $F$
(which is necessarily subsidiary to $A$) is {\em suitable}: its result
contains $tm$ as a subterm.  After the stack is printed, the result of
its final rewrite rule application is printed.  If the stack extends
beyond $A$ then the result of $A$ is also printed.

Next we give two examples that illustrate the description above.
Let's start with the simpler one.

{\footnotesize\begin{Verbatim}[commandchars=\\\{\}]
(include-book "std/lists/rev" :dir :system)
(with-brr-data
 (thm (implies (and (natp n)
                    (< n (len x)))
               (equal (nth n (revappend x y))
                      (nth n (reverse x))))
      :hints {\em{; The second example shows what happens when we remove :hints.}}
      (("Goal" :do-not '(preprocess)))))
\end{Verbatim}
}

\noindent We see the following checkpoint at the top level.

{\footnotesize\begin{verbatim}
(IMPLIES (AND (INTEGERP N)
              (<= 0 N)
              (< N (LEN X))
              (NOT (STRINGP X)))
         (EQUAL (NTH N (APPEND (REV X) Y))
                (NTH N (REV X))))
\end{verbatim}
}

\noindent At this point we might reasonably ask: How did rewriting
produce \texttt{(REV X)}?  Note that \texttt{REV} is not a built-in
function; it must have been defined in an included book.  The
following log, explained below, shows how a query using
\texttt{cw-gstack-for-subterm} can provide an answer.

{\footnotesize\begin{verbatim}
  ACL2 !>(cw-gstack-for-subterm (REV X))
  1. Simplifying the clause
       ((IMPLIES (IF (NATP N) (< N (LEN X)) 'NIL)
                 (EQUAL (NTH N (REVAPPEND X Y))
                        (NTH N (REVERSE X)))))
  2. Rewriting (to simplify) the atom of the first literal,
       (IMPLIES (IF (NATP N) (< N (LEN X)) 'NIL)
                (EQUAL (NTH N (REVAPPEND X Y))
                       (NTH N (REVERSE X)))),
  3. Rewriting (to simplify) the second argument,
       (EQUAL (NTH N (REVAPPEND X Y))
              (NTH N (REVERSE X))),
  4. Rewriting (to simplify) the first argument,
       (NTH N (REVAPPEND X Y)),
  5. Rewriting (to simplify) the second argument,
       (REVAPPEND X Y),
  6. Attempting to apply (:REWRITE REVAPPEND-REMOVAL) to
       (REVAPPEND X Y)
  The resulting (translated) term is
    (BINARY-APPEND (REV X) Y).
  ACL2 !>
\end{verbatim}
}

\noindent The use of \texttt{with-brr-data} above caused data to be
collected that we can query as shown in the log above.  Frame 1 shows
the initial \hrefdoc{CLAUSE}{clause} (list of literals, implicitly
disjoined), which in this case is a list containing just the initial
\hrefdoc{TERM}{translated} goal.  As we look down the stack we see how
the process of simplification moved from there to frame 6, which is
what the search was seeking: the first rewrite rule application $A$
that introduced \texttt{(REV X)} as a subterm; as noted above, we will
call $A$ the {\em product of the search}.  The stack stops there
  because no rewrite rules were applied to the right-hand side after
  applying \texttt{REVAPPEND-REMOVAL}.  The log concludes with the
  result of $A$.

The second example modifies the first by removing the \texttt{:hints}.
That produces the following proof output, for example by using
\hrefdoctt{PSO}{:pso}.

{\footnotesize\begin{verbatim}
By the simple :definition NATP and the simple :rewrite rule REVAPPEND-REMOVAL
we reduce the conjecture to

Goal'
(IMPLIES (AND (INTEGERP N)
              (<= 0 N)
              (< N (LEN X)))
         (EQUAL (NTH N (APPEND (REV X) Y))
                (NTH N (REVERSE X)))).
\end{verbatim}
}

\noindent This goal further simplified to produce the same checkpoint
as the first example.  But the word ``simple'' in the log above
indicates that simplification was performed with the lightweight
\hrefdoc{HINTS-AND-THE-WATERFALL}{``preprocess''} simplifier.  Like
break-rewrite, \texttt{with-brr-data} does not store data from the
preprocess simplifier.  In particular, nothing was stored while
generating the goal above.  Without such data, it is impossible to
track the source of the subterm \texttt{(REV X)} occurring in
\texttt{Goal'}, since it has already been put into that goal with
preprocessing and we are looking for rewrite rule applications that
{\em introduce} \texttt{(REV X)} as a subterm.

However, simplification of \texttt{Goal'} introduced a new occurrence
of \texttt{(REV X)}, as seen in the following log; explanation
follows.

{\footnotesize\begin{verbatim}
  ACL2 !>(cw-gstack-for-subterm (REV X))
  1. Simplifying the clause
       ((NOT (INTEGERP N))
        (< N '0)
        (NOT (< N (LEN X)))
        (EQUAL (NTH N (BINARY-APPEND (REV X) Y))
               (NTH N (REVERSE X))))
  2. Rewriting (to simplify) the atom of the fourth literal,
       (EQUAL (NTH N (BINARY-APPEND (REV X) Y))
              (NTH N (REVERSE X))),
  3. Rewriting (to simplify) the second argument,
       (NTH N (REVERSE X)),
  4. Rewriting (to simplify) the second argument,
       (REVERSE X),
  5. Attempting to apply (:DEFINITION REVERSE) to
       (REVERSE X)
  6. Rewriting (to simplify) the body,
       (IF (STRINGP X)
           (COERCE (REVAPPEND (COERCE X 'LIST) 'NIL)
                   'STRING)
         (REVAPPEND X 'NIL)),
     under the substitution
       X : X
  7. Rewriting (to simplify) the third argument,
       (REVAPPEND X 'NIL),
     under the substitution
       X : X
  8. Attempting to apply (:REWRITE REVAPPEND-REMOVAL) to
       (REVAPPEND X 'NIL)
  9. Rewriting (to simplify) the rhs of the conclusion,
       (BINARY-APPEND (REV X) Y),
     under the substitution
       Y : 'NIL
       X : X
  10. Attempting to apply (:REWRITE APPEND-ATOM-UNDER-LIST-EQUIV) to
       (BINARY-APPEND (REV X) 'NIL)
  The resulting (translated) term is
    (REV X).
  Note: The first lemma application above that provides a suitable result
  is at frame 5, and that result is
    (IF (STRINGP X)
        (COERCE (REV (COERCE X 'LIST)) 'STRING)
      (REV X)).
  ACL2 !>
\end{verbatim}
}

\noindent The clause in frame 1 corresponds to the goal above,
\texttt{Goal'}; thus this stack is from the process of simplifying
that goal.  Frames 2 and 3 show that we are dealing with the second
argument of the call of \texttt{EQUAL}, which unlike the first
argument did not already have a subterm of \texttt{(REV X)}.  Frame 5
shows the product of the search: the first rewrite rule application
that introduced \texttt{(REV X)} as a subterm, whose result is shown
in the Note printed at the end about ``a suitable result''.  For the
rest of the stack after frame 5, \texttt{(REV X)} is a subterm of the
result of each rewrite rule application.  Just after the stack is
printed (and before the Note at the end, which is about frame 5), we
see the result of the rule application of the final frame, frame 10,
which happens to be \texttt{(REV X)} itself.

The second example above illustrates why the stack is extended past
the product of the search using rule applications whose result
contains the subterm of the query, in this case \texttt{(REV X)}.  If
the display had ended at frame 5, we would not have seen the rule most
directly responsible for introducing \texttt{(REV X)} as a subterm ---
\texttt{REVAPPEND-REMOVAL}, applied at frame 8 -- even though
\texttt{(REV X)} occurs in the result from frame 5 (as noted at the
end of the log above).

For more examples, see the \hrefdoc{COMMUNITY-BOOKS}{community books}
file, \texttt{system/tests/brr-data-input.lsp} and corresponding log
\texttt{brr-data-log.txt} in that directory.

The following related query capabilities are also supported.

\begin{itemize}

\item The query \texttt{(cw-gstack-for-term $tm$)} differs from
  \texttt{(cw-gstack-for-subterm $tm$)} only in that it searches
  results for $tm$ itself, rather than for terms containing $tm$ as a
  subterm.

\item The queries \texttt{(cw-gstack-for-subterm* $tm$)} and
  \texttt{(cw-gstack-for-term* $tm$)} are {\em iterative} versions of
  their counterparts without the `\texttt{*}' suffix, in that that
  they query for additional results.  Each successive search in the
  scope of these queries ignores all rule applications that are
  subsidiary to the products of previous searches.

\item All of these utilities can take an argument of the following
  form (as permitted for \texttt{:expand} \hrefdoc{HINTS}{hints}):
  \texttt{(:free ($v_1$ $\ldots$ $v_k$) $tm$)}, where the $v_i$ are
  variables and $tm$ is a term.  In this case, the search is for any
  instance of $tm$ obtained by substituting for the $v_i$.  Note
  however that once the product of a search is found, the
  corresponding instance of $tm$ is used for finding a maximal stack
  extension.

\end{itemize}

We conclude this section with three remarks.

\begin{enumerate}

\item Rules are monitored within the scope of \texttt{with-brr-data}
  as though \texttt{:}\hrefdoctt{BRR}{brr}\texttt{ t} has been executed.

\item \texttt{With-brr-data} does not collect any data for the
  near-miss breaks discussed in the next section.

\item In \hrefdoc{PARALLELISM}{ACL2(p)}, \texttt{with-brr-data} is
  disallowed when
  \hrefdoc{SET-WATERFALL-PARALLELISM}{waterfall-parallelism} is
  enabled, as that would interfere with the sequential nature of data
  collection (which relies on the timing for collection of subsidiary
  applications).

\end{enumerate}

\section{User-level introduction to
  \hrefdoc{BREAK-REWRITE}{break-rewrite}}
\label{intro-to-brr}

\hrefdoc{BREAK-REWRITE}{Break-rewrite} was originally designed to help answer
the question ``why did the attempt to apply a certain lemma fail?''
It was modeled on Nqthm's \texttt{break-lemma}
\cite[pp. 257--264]{aclh}.\footnote{Nqthm's \texttt{break-lemma} is
  also described in \cite[pp. 305--311]{aclh2}.}
For example, suppose the user has proved these two rules,

{\footnotesize\begin{verbatim}
(defthm p-rule  (implies (q x) (p (f x y))))
(defthm q-rule1 (implies (r x) (q x)))
\end{verbatim}
}

{\noindent}and then tries \texttt{(thm (implies (r v) (p (f u v))))}.  The proof fails.
But the user expected the two rules to be used to prove the theorem and so responds with

{\footnotesize\begin{verbatim}
(brr t)                            ; turn on break-rewrite
(monitor 'p-rule  t)               ; unconditionally break when p-rule is matched
(monitor 'q-rule1 t)               ; unconditionally break when q-rule1 is matched
(thm (implies (r v) (p (f u v))))  ; try thm again
\end{verbatim}
}

This time there are interactive breaks.  The keyword commands below are the user's
responses to the breaks.  We have indented the depth 2 break for clarity; ACL2
does not indent breaks, to save space on the line.

{\footnotesize\begin{verbatim}
(1 Breaking (:REWRITE P-RULE) on (P (F U V)):
1 ACL2 >:eval

      (2 Breaking (:REWRITE Q-RULE1) on (Q U):
      2 ACL2 >:eval

      2x (:REWRITE Q-RULE1) failed because :HYP 1 rewrote to (R U).

      2 ACL2 >:type-alist

      Decoded type-alist:
      -----
      Terms with type (TS-COMPLEMENT *TS-NIL*):
      (R V)

      ==========
      Use (GET-BRR-LOCAL 'TYPE-ALIST STATE) to see actual type-alist.
      2 ACL2 >:a!
      Abort to ACL2 top-level.

\end{verbatim}
}

We see that \texttt{q-rule1} failed because its first hypothesis rewrote to \texttt{(R U)}
but the \hrefdoctt{TYPE-ALIST}{:type-alist} shows \texttt{(R V)} as a given.  We should be trying to prove
\texttt{(thm (implies (r u) (p (f u v))))}.

Break-rewrite was helpful here because the user expected a certain rule to be
tried and a target matching the rule's left-hand side was encountered by the
rewriter, but something prevented the application.  Typical failure reasons
are that a hypothesis could not be relieved, a suitable free variable
instantiation could not be found, or the rule would have performed a
``heuristically unattractive'' replacement.

But if the rewriter never sees a target that matches the rule, break-rewrite
cannot help us --- or at least it could not help us until the recent addition
of {\em{near-miss}} break criteria.

We will illustrate a near-miss break with a lemma about
\hrefdoctt{LOOP\_42}{loop\$}~\cite{acl2-loop}.
Near-miss break criteria are more general than this example might suggest.
We elaborate at the end of this section.  But four facts team up to make
lemmas about \texttt{loop\$} particularly prone to near-miss mismatches.  (a)
\texttt{Loop\$}s create quoted \hrefdoccltt{LAMBDA}{lambda} constants; (b) each
\texttt{lambda} constant contains arbitrary ACL2 code, namely the
\texttt{loop\$} body; (c) the prover can rewrite \texttt{lambda} constants in
slots of \hrefdoc{ILK}{ilk} \texttt{:FN}; but (d) matching requires identity
on quoted constants.  These facts often mean that lemmas about
\texttt{loop\$} fail to match because the \texttt{lambda} constants in the
lemmas are not in ``rewrite-normal form'' because we typically do not write
code in rewrite-normal form.  (``Rewrite-normal form'' is an informal notion.
A term is in rewrite-normal form if it is not changed by rewriting.  Of
course, this really depends on the context in which the term occurs.)

Here is an example.  Suppose \texttt{(nats n)} has been defined to return a
list of natural numbers and \texttt{foo} is some function of one argument.
Suppose the user wants to prove

{\footnotesize\begin{verbatim}
(defthm thm-a
  (loop$ for e in (nats (foo a)) always (atom e)))
\end{verbatim}
}
{\noindent}and has realized a more general lemma is needed:
{\footnotesize\begin{verbatim}
(defthm lemma-a
  (loop$ for e in (nats n) always (atom e)))
\end{verbatim}
}

This lemma is easily proved after including the standard
\texttt{"projects/apply/top"} book.  But when the user tries to prove
\texttt{thm-a} after the proving \texttt{lemma-a} the proof fails.
Monitoring \texttt{lemma-a} does not help: no break happens because no target
term matching the lemma ever arises.  What we need to do is see the targets
that are near-misses.

Because our lemma involves \texttt{loop\$} (and thus \texttt{lambda} constants)
we install a near-miss monitor that means ``break when a match fails due only
to mismatching \texttt{lambda} constants.''

{\footnotesize\begin{verbatim}
(monitor 'lemma-a '(:lambda t))
\end{verbatim}
}

Trying \texttt{thm-a} again produces an interactive break to which the user
types the \texttt{:lhs} command.

{\footnotesize\begin{verbatim}
(1 Breaking (:REWRITE LEMMA-A) on 
(ALWAYS$ '(LAMBDA (LOOP$-IVAR) (IF (CONSP LOOP$-IVAR) 'NIL 'T))
         (NATS (FOO A))):

The pattern in this rule failed to match the target.  However, this
is considered a NEAR MISS under the break criteria, 
(:CONDITION 'T :LAMBDA T), specified when this rule was monitored.
The following criterion is satisfied.

* :LHS matches :TARGET except at one or more quoted LAMBDA constants.

1 ACL2 >:lhs
(ALWAYS$ '(LAMBDA (LOOP$-IVAR) (ATOM LOOP$-IVAR))
         (NATS N))
\end{verbatim}
}

The \texttt{:target} is shown in the break header.  We see that the body of
the quoted \texttt{lambda} constant in \texttt{:target} is \texttt{(IF (CONSP
  LOOP\$-IVAR) 'NIL 'T)} but the body of the quoted \texttt{lambda} constant
in \texttt{:lhs} is \texttt{(ATOM LOOP\$-IVAR)}.  The \texttt{lambda}
constant in the theorem was rewritten; \texttt{(ATOM LOOP\$-IVAR)} was
expanded.

Problems of this sort can often be addressed in several ways, e.g., by
disabling the rewriting of \texttt{lambda} objects (see
\hrefdoc{REWRITE-LAMBDA-OBJECT}{rewrite-lambda-object}), or disabling the
functions in the body that the rewriter expanded.  But, as in this case,
those approaches often raise other issues.  It is generally best to restate
the lemma so that the body of the \texttt{lambda} is in rewrite-normal form.
That is, restate \texttt{LEMMA-A} so that the body of the \texttt{loop\$} is
\texttt{(if (consp e) nil t)} instead of \texttt{(atom e)}.  The experienced
ACL2 user would not formulate a rewrite rule containing a non-recursive
function like \texttt{atom} in its left-hand side but that is exactly what we
did in our original formulation of \texttt{lemma-a}.

Other near-miss criteria are \texttt{:depth k} where \texttt{k} is a natural
number and \texttt{:abstraction pat} where \texttt{pat} is a term.  The
former criterion triggers a break if the \texttt{:lhs} matches the target down
to depth \texttt{k}.  The latter triggers a break if the specified
\texttt{pat} matches the target.  All three criteria, \texttt{:lambda},
\texttt{:depth}, and \texttt{:abstraction}, are implemented the same way: the
near-miss criterion gives rise to a {\em{near-miss pattern}} that is typically
more general than the left-hand side, e.g., \texttt{:depth 2} applied to a
left-hand side of \texttt{(f (g (h x) x))} generates the near-miss pattern (F
(G GENSYM0 X)).  We then try to instantiate the near-miss pattern to produce
the target and if it succeeds we say that a near miss occurred (since
the left-hand side failed to unify), and a corresponding message is
printed.  Note that each near-miss criterion is handled this way, and
a message is printed for each that has occurred.
This process is carried out by the
function \hrefdoctt{BRR-NEAR-MISSP}{brr-near-missp}.  We hope to make
\texttt{brr-near-missp} attachable so users can add additional ways to trigger
near-miss breaks, but we have not done so yet.  See \hrefdoctt{MONITOR}{monitor}
and \hrefdoctt{BRR-NEAR-MISSP}{brr-near-missp} for details.

\section{Using \texttt{with-brr-data} and break-rewrite together}

Sometimes \texttt{with-brr-data} doesn't quite do the job by itself
but is useful in concert with break-rewrite, as illustrated by the
following example from the documentation for
\hrefdoctt{WITH-BRR-DATA}{with-brr-data}.  This example does not exercise
the near-miss feature of break-rewrite.

{\footnotesize\begin{verbatim}
(with-brr-data (thm (equal (append x y) (append y x))))
\end{verbatim}
}

\noindent After the proof attempt fails, the first checkpoint under
induction is as follows.

{\footnotesize\begin{verbatim}
(IMPLIES (AND (CONSP X)
              (EQUAL (APPEND (CDR X) Y)
                     (APPEND Y (CDR X))))
         (EQUAL (CONS (CAR X) (APPEND Y (CDR X)))
                (APPEND Y X)))
\end{verbatim}
}

\noindent Even an experienced user might be surprised, at least
initially, to see the second occurrence of \texttt{(APPEND Y (CDR
  X))}.  The following log shows how an appropriate query can attempt
to shed light on that.

{\footnotesize\begin{verbatim}
  ACL2 !>(cw-gstack-for-subterm (append y (cdr x)))
  1. Simplifying the clause
       ((NOT (CONSP X))
        (NOT (EQUAL (BINARY-APPEND (CDR X) Y)
                    (BINARY-APPEND Y (CDR X))))
        (EQUAL (BINARY-APPEND X Y)
               (BINARY-APPEND Y X)))
  2. Rewriting (to simplify) the atom of the third literal,
       (EQUAL (BINARY-APPEND X Y)
              (BINARY-APPEND Y X)),
  3. Rewriting (to simplify) the first argument,
       (BINARY-APPEND X Y),
  4. Attempting to apply (:DEFINITION BINARY-APPEND) to
       (BINARY-APPEND X Y)
  The resulting (translated) term is
    (CONS (CAR X)
          (BINARY-APPEND Y (CDR X))).
  ACL2 !>
\end{verbatim}
}

\noindent This shows us that the term \texttt{(BINARY-APPEND Y (CDR
  X))}, which is the translated \hrefdoc{TERM}{term} corresponding to
the user input of \texttt{(APPEND Y (CDR X))}, is produced from the
definition of \texttt{BINARY-APPEND}.  But how?  We can
\hrefdoc{MONITOR}{monitor} that definition to answer that question.
Here is a log, abbreviated as shown.

{\footnotesize\begin{Verbatim}[commandchars=\\\{\}]
  ACL2 !>:monitor! binary-append (equal (brr@ :target) '(BINARY-APPEND X Y))
   T
  ACL2 !>(thm (equal (append x y) (append y x)))

  {\em [[.. Use :go to get past the start of induction. ..]]}

  *1 (the initial Goal, a key checkpoint) is pushed for proof by induction.

  {\em [[.. elided ..]]}

  (1 Breaking (:DEFINITION BINARY-APPEND) on (BINARY-APPEND X Y):
  1 ACL2 >:eval

  1! (:DEFINITION BINARY-APPEND) produced 
  (CONS (CAR X) (BINARY-APPEND Y (CDR X))).

  1 ACL2 >:type-alist

  Decoded type-alist:
  -----
  Terms with type *TS-T*:
  (EQUAL (APPEND (CDR X) Y)
         (APPEND Y (CDR X)))
  -----
  Terms with type *TS-CONS*:
  X

  ==========
  Use (GET-BRR-LOCAL 'TYPE-ALIST STATE) to see actual type-alist.
  1 ACL2 >
\end{Verbatim}
}

\noindent We see that an equality --- in this case, the induction
hypothesis --- has been applied to switch the arguments of the call of
\texttt{BINARY-APPEND} that was created by applying its definition.

\section{Brief introduction to \hrefdoc{WORMHOLE}{wormholes}}

The implementations of \texttt{with-brr-data} and break-rewrite both
depend on ACL2 \hrefdoc{WORMHOLE}{wormhole}s.  Here we provide a bit
of relevant background on wormholes.

Every ACL2 function is a true mathematical function: equal inputs produce
equal outputs.  There are no side-effects.  To interact with the user a
function must take and return the ACL2 \hrefdoc{STATE}{state} so that
inputs can be read and outputs printed.

\hrefdoctt{WORMHOLE}{Wormhole} is an ACL2 function that takes a {\em{wormhole
    name}} and some other arguments, not including \hrefdoctt{STATE}{state},
and always returns \texttt{nil}.  So logically it is a trivial constant
function.  But when \texttt{wormhole} is called a new
read-eval-print loop is started on a copy of the ACL2 state which also
contains the inputs to the call of \texttt{wormhole} and an
object, called the {\em{wormhole status}} associated with wormhole name as of
the last time the wormhole was exited.  While in this loop, forms can print
information to the \hrefdoctt{CW}{comment window}, inspect and
compute a new wormhole status, and (to a limited extent) modify the copy of
the state available in the wormhole.  But the first form ``read'' and
executed by this loop is not one typed by the user but is provided in the
call of \texttt{wormhole}.  Thus the call of \texttt{wormhole} can inspect
the available information and configure the status as appropriate and then
either exit the loop (without ever prompting the user for anything) or stay
in the loop and prompt the user for input.  When the loop is exited for any
reason the final status is saved in a secret location to be reinstated the
next time that wormhole is entered.  The rest of that copy of the state is
discarded.

Thus, using a wormhole, you can accumulate into the status any data passed
into the wormhole or obtained from state, you can print data, and you can
interact with the user, but you cannot pass data from inside the wormhole out
to the caller: the result is always \texttt{nil} and the ACL2 state
remains unchanged.

Wormholes necessitate the ``copying'' of the ACL2 state and the saving of
data outside of the ACL2 state.  The copying is just an illusion.
The state inside a wormhole is the ``live state'' but all allowed changes are
tracked, including changes to \hrefdoc{PROGRAMMING-WITH-STATE}{state
  global variables},
and when the wormhole exits, those changes are undone by Lisp's unwind
protection mechanism.  (This actually messed up the behavior of break-rewrite,
e.g., by losing track of which lemmas are monitored,
when the user invoked the theorem prover recursively from within a wormhole; but
that problem has been fixed.)  As for saving data outside the state, we
use a raw Lisp association list to associate each wormhole name with its
current status.  When the wormhole is entered, that status object is assigned
to a state global variable in the
ACL2 state so that the status is visible to forms
executing in the wormhole.  When the wormhole is exited, that status is
written back to the raw Lisp association list.  This general scheme has been
in effect for 30 years, but in working on break-rewrite recently we
discovered some problems that necessitated clarifying the implementation of
wormholes.  The first step in that clarification is to introduce some
terminology: the status of a wormhole stored in raw Lisp is called the
{\em{persistent}} wormhole status (``whs''), while the status occasionally
stored in the ACL2 state is called the {\em{ephemeral}} whs because
it disappears and reappears.  The problem we discovered with the old
implementation of wormholes can best be understood by thinking of the
persistent whs as a hard-to-access memory location and the ephemeral one
as an easily accessed, nearby cache.  The problem was that our cache was not
always coherent: some functions (or user commands) changed one without
changing the other.  See \hrefdoctt{WORMHOLE}{wormhole} and
\hrefdoctt{WORMHOLE-PROGRAMMING-TIPS}{wormhole-programming-tips} for details.

Wormholes can be expensive to enter and exit, e.g., cleanup forms must be
consed up upon state changes and evaluated upon exit, and forms executed
within the wormhole must read, translated, and interpreted.  So we provide a
more efficient mechanism called \hrefdoctt{WORMHOLE-EVAL}{wormhole-eval},
which essentially takes a wormhole name and a \hrefdoccltt{LAMBDA}{lambda}
expression, binds the \texttt{lambda} formal to the persistent status of the
named wormhole, evaluates the term, and stores the result back into the
persistent status.

\section{Implementation aspects for
  \hrefdoc{BREAK-REWRITE}{break-rewrite}}

The ACL2 rewriter does not modify \hrefdoc{STATE}{state}, so break-rewrite is
implemented by writing to a wormhole named \texttt{brr}.  The status of the
\texttt{brr} wormhole is basically a state machine that records information
about the rewriter's activities.  The status is represented in an ACL2
\hrefdoc{DEFREC}{defrec} object.\footnote{This is a change made recently; for
the first 30 years of break-rewrite's implementation the status was factored
differently and when moved by \texttt{wormhole} from its persistent location
to its ephemeral location was scattered over four different state globals.
Now the status object is assigned to a single global.}

{\footnotesize\begin{verbatim}
(defrec brr-status
  (entry-code (brr-monitored-runes . brr-gstack)
              . (brr-local-alist . brr-previous-status))
  t)
\end{verbatim}
}

See \hrefdoctt{WORMHOLE}{wormhole} for an explanation of \texttt{entry-code}.
The next four components are the list of monitored runes and their break
criteria, the rewriter's call stack, an alist binding variables passed in
from the rewriter (and a few specific to the given break), and the previous
\texttt{brr-status}.  We think of a \texttt{brr-status} object as a stack:
\texttt{brr-monitored-runes}, \texttt{brr-gstack}, and
\texttt{brr-local-alist} characterize an active (still open) call of
break-rewrite and \texttt{brr-previous-status} is the stack of calls leading
to this one.

We have sprinkled calls of three {\em{breakpoint handlers}} throughout the
mutually recursive clique of 52 functions implementing the ACL2 rewriter.
These functions do nothing until break-rewrite is turned on with \texttt{(brr
  t)} or within the scope of \texttt{with-brr-data}.  Logically the breakpoint handlers are no-ops that return
\texttt{nil}.  But when \texttt{(brr t)} has been done, a handler
may enter a \texttt{brr} wormhole to save or erase data and to interact with the user.
\begin{itemize}

\item \texttt{Near-miss-brkpt1} is called when the rewriter considers a
  rule but finds that the rule's left-hand side fails to match the current
  target.  The \texttt{brr} wormhole is entered.  The first thing that
  happens inside the \texttt{brr} wormhole when invoked by this handler is to
  determine whether the rule being considered by the rewriter is monitored
  and has near-miss criteria associated with it.  These questions can only be
  answered from inside the wormhole since the rewriter has no information
  about monitored rules.  If the answers are affirmative, the target
  is compared to the near-miss
  pattern of each specified near-miss criterion.  If any near-miss pattern
  matches the target, the function pushes a new status on the stack of
  statuses, prints an ``open break banner'' explaining each of the
  near-misses just detected, and prompts the user for input.
  When the user issues the command to
  proceed from the break, the wormhole and \texttt{near-miss-brkpt1} are
  exited.  The rewriter continues as it would had the handler never been
  called.  It will, in fact, subsequently call \texttt{brkpt2} discussed
  below on the very same rewrite call stack.  That will allow \texttt{brkpt2}
  to detect that it should print a ``close break banner'' and pop the
  \texttt{brr-status} stack.

\item \texttt{Brkpt1} is called when the rewriter considers a rule and finds
  that the rule's left-hand side matches the current target.  The
  \texttt{brr} wormhole is entered.  If the rule is monitored and the break
  condition is satisfied, the function pushes a new status on the stack,
  prints an open break banner explaining the break, and prompts the user for
  input.  When the user types an exit command the \texttt{brr-status} is
  updated to record which exit command was used.  The
  \hrefdoc{BRR-COMMANDS}{\texttt{:eval} command} means proceed to try to
  apply the rule and reenter this interactive break when the attempt is
  complete, \texttt{:go} means proceed, print the results of the attempt but
  do not interact further with the user, and \texttt{:ok} means proceed and
  do not even print the results.  There are variants for controlling whether
  further breaks are allowed while attempting to apply the current rule.  In
  all cases, these commands cause the wormhole and \texttt{brkpt1} to
  exit.  The rewriter proceeds as though \texttt{brkpt1} had never been
  called, trying to relieve the hypotheses, test heuristic conditions, etc.
  It will eventually call \texttt{brkpt2} on the same call stack.

\item \texttt{Brkpt2} is called when the rewriter is finished considering a
  rule.  The \texttt{brr} wormhole is entered, with
  information passed in from the rewriter that includes the rewriter's call
  stack and data about what happened, e.g., whether the attempt succeeded or
  not, if not, why not, etc.  If the rewriter's call stack is the same as the
  \texttt{brr-gstack} in the current \texttt{brr-status}, then we know this is
  the balancing ``closing'' phase of that open break.  In this case,
  \texttt{brkpt2} either prompts the user for more input (if the opening
  break was exited with \texttt{:eval}) or just prints the appropriate close
  break banner, pops the \texttt{brr-status}, and exits the wormhole.  In the
  case that \texttt{brkpt2} prompts the user for input then the closing
  banner, the stack pop, and exit happen when the user issues an exit
  command.

\end{itemize}

Thus each \texttt{near-miss-brkpt1} call that opened a break is balanced by a
\texttt{brkpt2} call that closes it, and each \texttt{brkpt1} call that
opened a break is balanced by a \texttt{brkpt2} call that closes
it.\footnote{Care is taken to clean up the \texttt{brr-status} stack in the
event of an error exit or interrupt.}  But there can be additional
\texttt{near-miss-brkpt1}, \texttt{brkpt1} and \texttt{brkpt2} calls between
a balanced pair that signals breaks, all from rewrite rule applications
that are subsidiary to the one handled by that balanced pair, as we
now illustrate.

Suppose we are in the state discussed in Section \ref{intro-to-brr} where we
had the first two rules noted below, but add the third rule.

{\footnotesize\begin{verbatim}
(defthm p-rule  (implies (q x) (p (f x y))))
(defthm q-rule1 (implies (r x) (q x)))
(defthm q-rule2 (implies (s x) (q x)))
\end{verbatim}
}

Monitor the first two rules as before, issue the \texttt{thm} command
below, and type \texttt{:GO} to every break.  The left column below shows all
calls of \texttt{brkpt1} and \texttt{brkpt2} as obtained by tracing those two
functions and just printing their names.  We have further annotated those
calls with the name, in brackets, of the lemma being considered by the rewriter
when the breakpoint handler is called.  Thus, ``\texttt{> BRKPT1
  \{p-rule\}}'' means we enter \texttt{brkpt1} with \texttt{p-rule} being
considered by the rewriter, and ``\texttt{< BRKPT1 \{p-rule\}}'' means we
exit \texttt{brkpt1} with \texttt{p-rule} being considered. The right column
shows the output of break-rewrite and the user's responses.  We have indented
the break output and deleted some blank lines.

{\footnotesize\begin{verbatim}
ACL2 !>(thm (implies (r u) (p (f u v))))
1> BRKPT1 {p-rule}
                            (1 Breaking (:REWRITE P-RULE) on (P (F U V)):
                            1 ACL2 >:GO
<1 BRKPT1 {p-rule}
1> BRKPT1 {q-rule2}
<1 BRKPT1 {q-rule2}
1> BRKPT2 {q-rule2}
<1 BRKPT2 {q-rule2}
1> BRKPT1 {q-rule1}
                                  (2 Breaking (:REWRITE Q-RULE1) on (Q U):
                                  2 ACL2 >:GO
<1 BRKPT1 {q-rule1}
1> BRKPT2 {q-rule1}
                                  2 (:REWRITE Q-RULE1) produced 'T.
                                  2)
<1 BRKPT2 {q-rule1}
1> BRKPT2 {p-rule}
                            1 (:REWRITE P-RULE) produced 'T.
                            1)
<1 BRKPT2 {p-rule}
Q.E.D.
\end{verbatim}
}

It might be surprising that each trace is at level 1: no call of
\texttt{brkpt1} or \texttt{brkpt2} is within any other such call, even though
the apparent calls of break-rewrite are nested.  Break-rewrite is an illusion.
No such function is defined in ACL2 (which is why we never write it in typewriter
font).

The break-rewrite {\em{depths}} printed in the banners and prompts
correspond to the
\texttt{brr\--previous\--status} chain (via the brr stack depth). The break by
\texttt{brkpt1} at depth 1, when the rewriter is considering \texttt{p-rule},
pushes a new status on the \texttt{brr-status} stack, prints a banner opening
the break, reads the user's \texttt{:go}, and exits.  The rewriter proceeds
to try to relieve the hypothesis, \texttt{(q u)}, of \texttt{p-rule}.  First
it tries \texttt{q-rule2}.  But when \texttt{brkpt1} is called on the
unmonitored \texttt{q-rule2}, \texttt{brkpt1} just exits silently.  The
rewriter fails to relieve the hypothesis of \texttt{q-rule2} and calls
\texttt{brkpt2} on \texttt{q-rule2}, which exits silently since the
rewriter's call stack is not the \texttt{brr-gstack} of the status.  Next the
rewriter tries \texttt{q-rule1}, calling \texttt{brkpt1} on \texttt{q-rule1},
which is monitored.  \texttt{Brkpt1} pushes another status on the
\texttt{brr-status} stack making the depth 2.  The \texttt{:go} at depth 2
exits that \texttt{brkpt1} and allows the rewriter to proceed to successfully
establish the hypotheses of \texttt{q-rule1} and then call \texttt{brkpt2}.
It detects that the rewriter's call stack is the \texttt{brr-gstack} of the
status and that the balancing \texttt{brkpt1} at depth 2 proceeded with
\texttt{:go}, so \texttt{brkpt2} prints the results, does not prompt for user
input, prints the closing banner for depth 2, pops the \texttt{brr-status}
stack, and exits.  The same thing happens (at depth 1) when \texttt{brkpt2} is eventually
called on \texttt{p-rule}.

\section{Implementation aspects for \hrefdoctt{WITH-BRR-DATA}{with-brr-data}}

We have seen that \hrefdoctt{WITH-BRR-DATA}{with-brr-data} supports
{\em saving} of relevant data during a proof attempt, to be used by
tools for {\em querying} the data.  In this section we focus primarily on
{\em saving} data, concluding with a few words about {\em
  querying} data.

Our approach to {\em saving} data is based on the observation that
calls of \texttt{brkpt1} and \texttt{brkpt2} are in exactly the places
we want to consider: before and after each matched rule is considered.
As noted earlier, \texttt{with-brr-data} collects no data for
near-miss breaks; thus, here we consider only \texttt{brkpt2} calls
that balance \texttt{brkpt1} calls rather than balancing
\texttt{near-miss-brkpt1} calls.  (Technically, we restrict to
those \texttt{brkpt2} calls for which the \texttt{failure-reason} argument
is not the symbol, \texttt{near-miss}.)  Note that unlike
break-rewrite, there is no need for \texttt{brkpt2} to check the
rewriter's call stack to check for balancing with a \texttt{brkpt1}
call, since there is data collection for every \texttt{brkpt1} call.

Thus, \texttt{with-brr-data} piggybacks on \texttt{brr} in the sense that
we modified those two breakpoint handlers to collect data when
appropriate, to be queried later.  But recall the focus on the source
of a term in a {\em checkpoint}.  Therefore we consider only top-level
rewrite rule applications, that is, we ignore rewrites that take place
during backchaining.  The functions in the ACL2 rewriter take an
\texttt{ancestors} argument that is non-\nil precisely during
backchaining, so we restrict data collection to when
\texttt{ancestors} is \nil.

But how can we store global data when the rewriter does not modify the ACL2
\hrefdoc{STATE}{state}?  As for break-rewrite, we use a
\hrefdoc{WORMHOLE}{wormhole} state.  We initially did this by modifying the
\texttt{brr} wormhole status, but that resulted in very slow execution
because, unlike typical uses of break-rewrite, \texttt{with-brr-data} saves
data at {\em every} \texttt{brkpt1} and \texttt{brkpt2} call for which
\texttt{ancestors} is \nil.  We solved this problem by making separate
\hrefdoc{WORMHOLE-EVAL}{wormhole-eval} calls to save data for
\texttt{with-brr-data} into a different wormhole state, named
\texttt{brr-data}.

That said, we still prefer to avoid calling even
\texttt{wormhole-eval} when no data is to be stored.  Here is the
relevant code in \texttt{brpkt1}, with the \texttt{wormhole-eval} call
abbreviated; the code in \texttt{brpkt2} is the same except that it
uses \texttt{brkpt2-brr-data-entry} instead of
\texttt{brkpt1-brr-data-entry}.

{\footnotesize\begin{verbatim}
(and (eq gstackp :brr-data)
     (brkpt1-brr-data-entry ancestors gstack rcnst state)
     (wormhole-eval 'brr-data ...))
\end{verbatim}
}

\noindent The first test is true in the scope of
\texttt{with-brr-data}, as we'll discuss later.  The second test is
true when \texttt{ancestors} is \nil (but the next section discusses
how that can be changed).  Only when those two conditions are met do
we call \texttt{wormhole-eval} to store appropriate data in the
\texttt{brr-data} wormhole.

The \texttt{wormhole-eval} call invokes functions to update the
\texttt{brr-data} wormhole state: \texttt{update\--brr\--data-1} in
\texttt{brkpt1} and \texttt{update\--brr\--data-2} in \texttt{brkpt2}.
More precisely, the wormhole state consists of a list of
\texttt{brr-data} records, which we now describe, and these two
functions update that list.

A \texttt{brr-data} record contains fields \texttt{pre} and
\texttt{post} that are \texttt{brr-data-1} and \texttt{brr-data-2}
records, respectively; see below.  \texttt{Pre} and \texttt{post}
contain information from balanced \texttt{brkpt1} and \texttt{brkpt2}
calls.  A \texttt{brr-data} record also has a \texttt{completed} field,
which is a list of \texttt{brr-data} records representing subsidiary
rewrite rule applications (as further described below).

{\footnotesize\begin{verbatim}
(defrec brr-data
  (pre post . completed)
  nil)

(defrec brr-data-1
  (((lemma . target) . (unify-subst . type-alist))
   .
   ((pot-list . ancestors) . (rcnst initial-ttree . gstack)))
  nil)

(defrec brr-data-2
  ((failure-reason unify-subst . brr-result)
   .
   (rcnst final-ttree . gstack))
  nil)
\end{verbatim}
}

\noindent The \texttt{completed} field of a \texttt{brr-data} record $B$
is a list of \texttt{brr-data} records created for balanced pairs of
\texttt{brkpt1}/\texttt{brkpt2} calls that took place between
\texttt{pre} and \texttt{post} fields, hence for the rewrite rule
applications subsidiary to the one represented by $B$.  Consider for
example a rewrite rule $r_1$, \texttt{(equal (f1 x) (f2 x))}, and a
rewrite rule $r_2$, \texttt{(equal (f2 x) (f3 x))}.
So the application of rule $r_1$ to
\texttt{(f1 a)} would generate an application of $r_2$ to \texttt{(f2
  x)} with \texttt{x} bound to \texttt{a}, so that \texttt{(f3 a)} is
the result of applying $r_2$ and hence of applying $r_1$ as well.
This process would be recorded in a \texttt{brr-data}
record whose \texttt{pre} field would have a
\texttt{target} of \texttt{(f1 a)} and whose \texttt{post} field would
have a \texttt{brr-result} field of \texttt{(f3 a)}.  Its
\texttt{completed} field would have a single \texttt{brr-data} record
representing the subsidiary application of rule $r_2$, hence with
\texttt{pre} and \texttt{post} fields whose \texttt{target} and
\texttt{brr-result} fields are \texttt{(f2 a)} and \texttt{(f3 a)},
respectively.

The \texttt{brr-data-1} and \texttt{brr-data-2} structures include
considerable information that isn't used by the built-in query
utilities.  However, \texttt{with-brr-data} doesn't cause much
slowdown, and the extra data, made readily available by the formal
parameters of \texttt{brkpt1} and \texttt{brkpt2}, may be useful for
user-defined attachments as discussed in the next section.

\texttt{With-brr-data} sets things up as follows to create
\texttt{brr-data} records, as required for queries like
\texttt{cw-gstack-for-subterm}.

\begin{enumerate}

\item Evaluate \texttt{(clear-brr-data-lst)} to remove previously
  saved data.

\item Assign state global \texttt{gstackp} to have
  value \texttt{:brr-data}.  Note that in the test \texttt{(eq
    gstackp :brr-data)} above, the variable \texttt{gstackp} is the
  value of that state global.

\item Evaluate the argument of \texttt{with-brr-data}, to invoke the
  prover.

\item Set state global \texttt{brr-data-lst} based on the stored data,
  by calling \texttt{(brr-data-lst state)}.

\end{enumerate}

\noindent The last step is interesting in a couple of ways.  First,
note that with state global \texttt{gstackp} set to
\texttt{:brr-data}, the prover populates the \texttt{brr-data} data
wormhole state with a list of \texttt{brr-data} records, one for each
rewrite rule application that is not subsidiary to any other
(generally from rewriting a literal of a \hrefdoc{CLAUSE}{clause}).
The list is constructed in reverse order: as the proof proceeds, new
records are pushed onto the front of that list.  What's more, each
completed field of each brr-data record is similarly in reverse order.
So the last step above puts everything into the right order before
storing the \texttt{brr-data} wormhole data into the state global,
\texttt{brr-data-lst}.  It may seem odd logically to obtain data from
the \texttt{brr-data} wormhole outside that wormhole.  The function
\hrefdoctt{GET-PERSISTENT-WHS}{get-persistent-whs} provides the
logical explanation by obtaining such data by
\hrefdoc{READ-ACL2-ORACLE}{reading the \texttt{acl2-oracle} field} of
the ACL2 \hrefdoc{STATE}{state}, which changes the state --- though
raw Lisp code for \texttt{get-persistent-whs} gets the result from the
persistent wormhole status of the \texttt{brr-data} wormhole.

We conclude this section by commenting briefly on the implementation
of the query utilities.  These traverse the state global described
above, \texttt{brr-data-lst}, searching for a \texttt{brr-data} record
that represents a rewrite rule application introducing the specified
subterm or term (or instance, in the \texttt{:free} case).  The source
code definitions of \texttt{cw-gstack-for-subterm},
\texttt{cw-gstack-for-term}, as well as their iterative (`\texttt{*}')
versions, are all reasonably straightforward.  Also see the
\hrefdoc{WITH-BRR-DATA}{documentation for \texttt{with-brr-data}} for
discussion of its keyword arguments and a few more
implementation-level details.

\section{Changing the behavior of \texttt{with-brr-data}}

The preceding section mentions functions \texttt{update-brr-data-1}
and \texttt{update-brr-data-2}, which are invoked in \texttt{brkpt1}
and \texttt{brkpt2}, respectively, to update the list of
\texttt{brr-data} records held in the \texttt{brr-data} wormhole
state.  These two functions are actually stubs that have respective
\hrefdoc{DEFATTACH}{attachments}
\texttt{update\--brr\--data\--1\--builtin} and
\texttt{update\--brr\--data\--2\--builtin}, which implement the steps
enumerated in the preceding section in a reasonably straightforward
way.

In fact, \texttt{with-brr-data} was originally designed and
implemented for collecting failed attempts at backchaining, rather
than for collecting appropriate top-level rewrites as is done now.
That original functionality is available by changing those attachments
after including the community book,
\texttt{kestrel/\-utilities/\-brr\--data\--failures.lisp}, and then
issuing a single command.  Below is that command and its single-step
macroexpansion.  It shows how the argument provided to the command, in
this case \texttt{failures}, provides a suffix for the attached
function names.

{\footnotesize\begin{verbatim}
ACL2 !>:trans1 (set-brr-data-attachments failures) ; whitespace edited below
 (WITH-OUTPUT :OFF :ALL
     (PROGN (DEFATTACH (UPDATE-BRR-DATA-1 UPDATE-BRR-DATA-1-FAILURES)
                       :SYSTEM-OK T)
            (DEFATTACH (UPDATE-BRR-DATA-2 UPDATE-BRR-DATA-2-FAILURES)
                       :SYSTEM-OK T)
            (DEFATTACH (BRKPT1-BRR-DATA-ENTRY BRKPT1-BRR-DATA-ENTRY-FAILURES)
                       :SYSTEM-OK T)
            (DEFATTACH (BRKPT2-BRR-DATA-ENTRY BRKPT2-BRR-DATA-ENTRY-FAILURES)
                       :SYSTEM-OK T)))
ACL2 !>
\end{verbatim}
}

The community book \texttt{kestrel/utilities/brr-data-all.lisp} is
similar except that it arranges to collect data for all rewrites, not
just for failed backchaining.  Just as the suffix
``\texttt{failures}'' was used above, the suffix ``\texttt{all}'' is
used for this ``\texttt{-all}'' book.  After including it, one would
evaluate \texttt{(set-brr-data-attachments all)} to get the desired
behavior via attachments.

Other behaviors can be implemented similarly.  To take advantage of
any of these, however, one might want to write suitable query
utilities, perhaps modeled on the implementation of the existing query
utilities such as \texttt{cw-gstack-for-subterm}.

\section{Conclusion}

Both \hrefdoctt{WITH-BRR-DATA}{with-brr-data} and
\hrefdoc{BREAK-REWRITE}{break-rewrite} can be very useful tools in
proof debugging.  We have shown how to use them and we have given a
glimpse of implementation issues and solutions.

\textbf{Acknowledgments.}  This research was developed with funding
from the Defense Advanced Research Projects Agency (DARPA).  The
views, opinions and/or findings expressed are those of the authors and
should not be interpreted as representing the official views or
policies of the Department of Defense or the U.S. Government.
We also thank ForrestHunt, Inc. for supporting research reported
herein, as well as the reviewers for helpful feedback.

\bibliographystyle{eptcs}
\bibliography{brr}

\begin{thebibliography}{1}
\providecommand{\bibitemdeclare}[2]{}
\providecommand{\surnamestart}{}
\providecommand{\surnameend}{}
\providecommand{\urlprefix}{Available at }
\providecommand{\url}[1]{\texttt{#1}}
\providecommand{\href}[2]{\texttt{#2}}
\providecommand{\urlalt}[2]{\href{#1}{#2}}
\providecommand{\doi}[1]{doi:\urlalt{https://doi.org/#1}{#1}}
\providecommand{\eprint}[1]{arXiv:\urlalt{https://arxiv.org/abs/#1}{#1}}
\providecommand{\bibinfo}[2]{#2}

\bibitemdeclare{book}{aclh}
\bibitem{aclh}
\bibinfo{author}{R.~S. \surnamestart Boyer\surnameend} \& \bibinfo{author}{J~S.
  \surnamestart Moore\surnameend} (\bibinfo{year}{1988}):
  \emph{\bibinfo{title}{A Computational Logic Handbook}}.
\newblock \bibinfo{publisher}{Academic Press}, \bibinfo{address}{New York}.

\bibitemdeclare{book}{aclh2}
\bibitem{aclh2}
\bibinfo{author}{R.~S. \surnamestart Boyer\surnameend} \& \bibinfo{author}{J~S.
  \surnamestart Moore\surnameend} (\bibinfo{year}{1997}):
  \emph{\bibinfo{title}{A Computational Logic Handbook, Second Edition}}.
\newblock \bibinfo{publisher}{Academic Press}, \bibinfo{address}{New York}.

\bibitemdeclare{book}{acl2-plus-books-manual}
\bibitem{acl2-plus-books-manual}
\bibinfo{author}{M.~\surnamestart Kaufmann\surnameend},
  \bibinfo{author}{\surnamestart \mbox{J} S.~Moore\surnameend} \&
  \bibinfo{author}{\surnamestart \mbox{The ACL2 Community}\surnameend}
  (\bibinfo{year}{2021}): \emph{\bibinfo{title}{The Combined {ACL2+Books}
  User's Manual}}.
\newblock \bibinfo{publisher}{\url{http://acl2.org/manual/index.html}}.

\bibitemdeclare{article}{acl2-loop}
\bibitem{acl2-loop}
\bibinfo{author}{Matt \surnamestart Kaufmann\surnameend} \&
  \bibinfo{author}{J~Strother \surnamestart Moore\surnameend}
  (\bibinfo{year}{2020}): \emph{\bibinfo{title}{Iteration in ACL2}}.
\newblock {\slshape \bibinfo{journal}{Electronic Proceedings in Theoretical
  Computer Science}} \bibinfo{volume}{327}, p. \bibinfo{pages}{16–31},
  \doi{10.4204/eptcs.327.2}.

\end{thebibliography}
\end{document}